\documentclass[letterpaper, 10 pt, conference]{ieeeconf}  

\IEEEoverridecommandlockouts                              

\usepackage[colorlinks=true, linkcolor=blue, urlcolor=red, citecolor=green]{hyperref}
\usepackage{soul}
\usepackage{cite}
\usepackage{subfig}
\usepackage{amsmath,amssymb,amsfonts}
\usepackage{graphicx}
\usepackage{booktabs}
\usepackage{multirow}
\usepackage{enumerate}
\usepackage{amsfonts}
\usepackage{subcaption}
\usepackage[table,xcdraw]{xcolor}

\usepackage{amsmath}
\usepackage{amssymb}
\usepackage{algorithm}
\usepackage{algpseudocode}

\usepackage{adjustbox}

\DeclareMathOperator*{\argmax}{arg\,max}

\title{\LARGE \bf
ActiveReg: Information-Driven Active Regional Probing for Partial-to-Full Bone Registration
}

\author{Tiancheng Li$^{1}$, Yingyu Wang$^{2}$, Peter Walker$^{3}$, Liang Zhao$^{2}$, and Shoudong Huang$^{1}$   
\thanks{This work was supported by the  National Health and Medical Research Council (NHMRC) Ideas Grant, Australia (No. 2029811).}
\thanks{$^{1}$Tiancheng Li and Shoudong Huang are with the Robotics Institute, Faculty of Engineering and Information Technology, University of Technology Sydney (UTS), Australia.}%
\thanks{$^{2}$Yingyu Wang and Liang Zhao are with the School of Informatics, The University of Edinburgh, United Kingdom.}%
\thanks{$^{3}$Peter Walker is with the Concord Repatriation General Hospital, New South Wales, Australia.}%
}

\begin{document}

\maketitle
\thispagestyle{empty}
\pagestyle{empty}

\begin{abstract}

Accurate bone registration is essential for orthopedic surgical navigation and robotic assistance. In a common workflow, the surgeon needs to identify and probe a number of prescribed locations on the exposed bone surface to obtain data points for registration, which can be difficult and time-consuming under limited surgical exposure. Reducing the number of required points while providing clear probing guidance would ease the surgeon's acquisition task intra-operatively. We present ActiveReg, a closed-loop framework that recommends probing regions rather than individual points, allowing flexibility in the exact contact location. A D-optimal planner that takes into account the already acquired points and the uncertainty of current registration information helps achieve accurate registration with fewer points. An online assessment combines pre-update innovation consistency with uncertainty at predefined surgical task locations improves the reliability of completion decision, without requiring ground truth. Simulations across four anatomical scenarios demonstrated that ActiveReg can achieve competitive accuracy compared with baseline registration methods while using substantially fewer acquired points. Real phantom experiments with optical and electromagnetic tracking demonstrated comparable mean target registration error to the Gradient-SDF framework, using only 24-28 acquired points instead of 771-1263.


\end{abstract}

\section{INTRODUCTION}
\label{introduction}

Surgical navigation and robotic-assisted technologies in orthopedic surgery require pre-operative plans to be accurately mapped to the patient’s anatomy \cite{digioia1998computer}. In total hip and knee arthroplasty, this mapping supports bone preparation and implant placement \cite{lonner2019robotics,li2024robotic,wang2025physical}. Bone registration achieves this mapping by establishing the spatial relationship between the pre-operative model and intra-operative patient coordinate frame. In a common workflow, intra-operative points are acquired from the exposed bone surface using a tracked probe and registered to a pre-operative model derived from computed tomography (CT) \cite{min20233d}. Reducing the number of probe contacts can ease the surgeon's acquisition effort, provided that the collected points retain sufficient geometric constraints for accurate registration. The acquisition strategy must therefore address both the informativeness of the points to be acquired how surgeons are guided during probing.

\begin{figure}[t]
	\centering
	\includegraphics[width=1\linewidth]{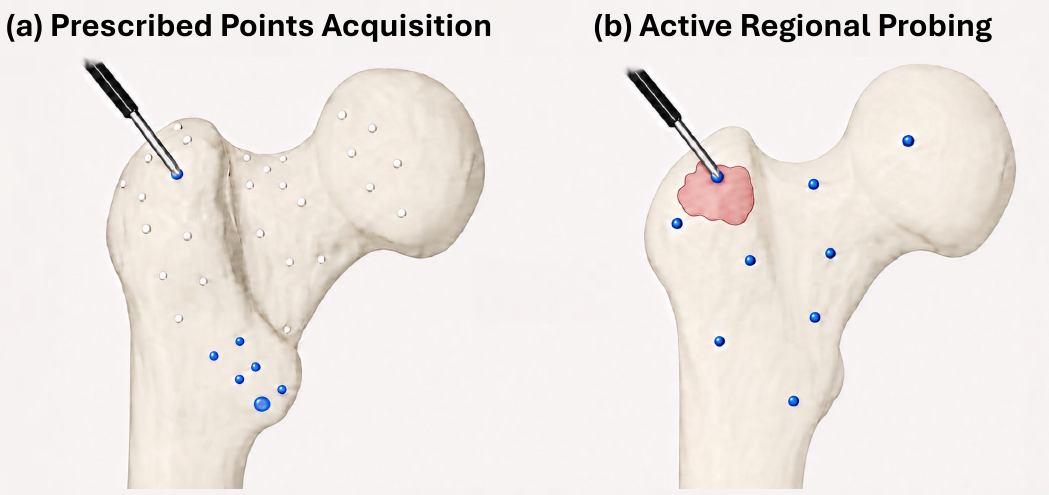}
	\caption{Comparison of two methods for probing guidance. Blue dots indicate already acquired points. (a) Commonly used prescribed points acquisition (MAKO \cite {perazzini2020mako}) : White dots mark prescribed points yet to be acquired, requiring precise probe placement. (b) Proposed active regional probing (ActiveReg): The red region represents the recommended area for the next probing location, allowing flexible contact placement and much fewer acquired points.}
	\label{fig:Intro}
\end{figure}

One approach is to guide intra-operative probing using prescribe surface points. For example, the current state-of-the-art orthopedic robotic system, MAKO Robotic Arm Interactive Orthopedic System (Stryker), specifies three initial anatomical landmarks followed by 32 prescribed points for each of the acetabulum and femur in the enhanced workflow \cite{perazzini2020mako}. These instructions requires the surgeon to identify and accurately contact the prescribe locations under limited exposure (Fig. \ref{fig:Intro}(a)), which becomes even more difficult when the bone surface is obscured by blood and surrounding tissue \cite{min2020generalized}. Correspondence-free surface registration methods offer greater flexibility in contact placement since acquired points do not need to be paired with prescribed model points. For example, the Gradient-SDF framework \cite{li2025partial} supports bone registration from sparse points freely acquired with a surgical probe moving on the exposed bone surface, without requiring predefined point correspondences. However, this freedom leaves the surgeon to decide where to probe and requires many contacts to acquire.

This motivates an acquisition strategy that provides explicit guidance on where to place the probe while allowing flexibility in the exact contact location. A straightforward way is to recommend a surface region within which the surgeon can choose where to probe, without having to identify and contact a single prescribed point precisely (Fig. \ref{fig:Intro}(b)). To reduce the number of contacts, the recommended region should guide the acquisition toward geometric information that complements the constraints provided by previously acquired contacts \cite{gelfand2003geometrically}. Since the surgeon can choose the contact location within each region, subsequent recommendations should be updated using the actual acquired points to address the remaining registration uncertainty. Alongside guidance on where to probe, an online completion assessment determines whether the acquired measurements support the required registration accuracy or further probing is needed.

\begin{figure*}[t]
	\centering
	\includegraphics[width=1\linewidth]{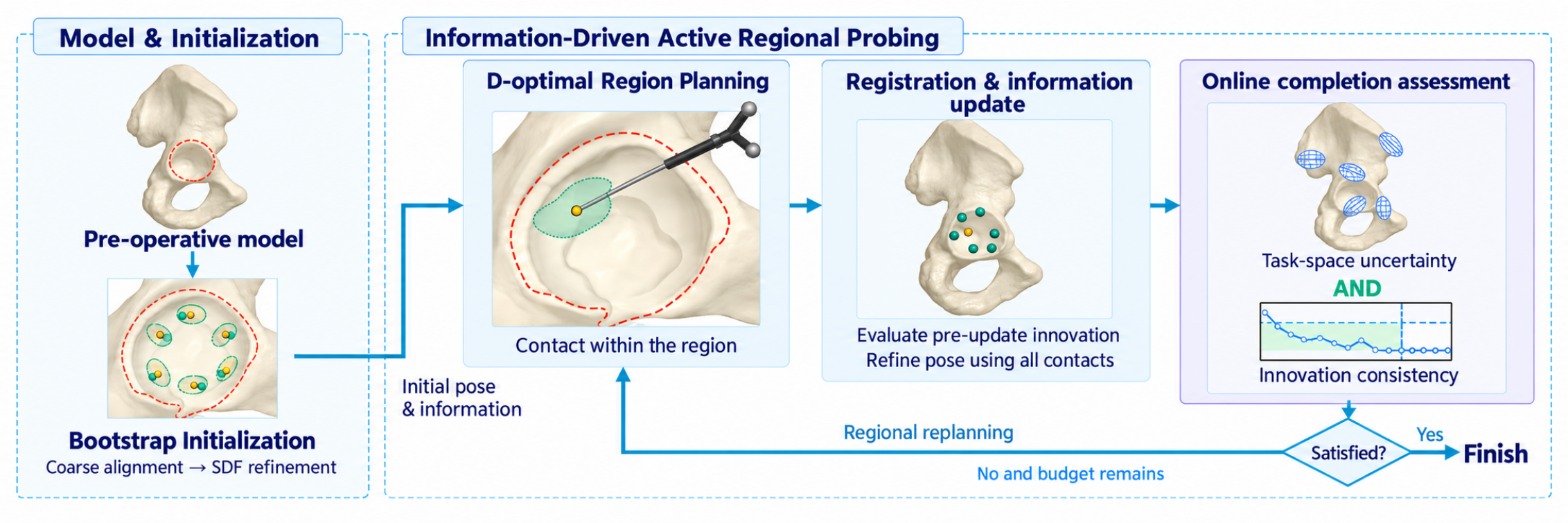}
	\caption{ActiveReg's closed-loop workflow. Six points acquired from bootstrap regions initialize the registration estimate and information matrix. D-optimal planning then recommends a region for the next probe contact. The actual point acquired within this region is used to update the registration estimate and information matrix. Acquisition continues until the joint innovation-consistency and task-space uncertainty criterion are satisfied or the point budget is exhausted.
    }
	\label{fig:Framework}
\end{figure*}

Therefore, we present ActiveReg, a closed-loop framework for achieving accurate bone registration with few probe contacts through
information-driven active regional probing and online completion assessment. ActiveReg uses the current registration estimate and information matrix to plan an informative surface region for the next probe contact. The surgeon probes a location within this region, and the actual acquired point is incorporated into correspondence-free signed distance field (SDF) registration to update the pose estimate and information matrix. Online completion assessment then determines whether to end acquisition or use the updated registration state to plan the next region.

The main contributions are: 
\begin{enumerate}
    \item  A closed-loop active regional probing framework that uses the current registration estimate to guide acquisition and uses actual probe measurements to update registration, reducing point requirements while maintaining registration accuracy. 

    \item A D-optimal region construction combining information gain with connectivity and minimum-area requirements, providing explicit guidance while allowing flexible contact placement within the recommended region.

    \item An online completion assessment that combines pre-update innovation consistency with task-space uncertainty to determine whether further probing is needed.
\end{enumerate}

\section{Related Work}
\label{sec:related work}

\subsection{Registration for Orthopedic Surgery}

Registration is a fundamental task to estimate a rigid transformation that align the geometric datasets. The Iterative Closest Point (ICP) algorithm \cite{besl1992method} is considered a milestone in point cloud registration, which alternates between closest-point correspondence estimation and transformation updates. Coherent point drift (CPD) \cite{myronenko2010point} formulates alignment as a Gaussian mixture model estimation problem. Go-ICP \cite{yang2015go} combines branch-and-bound search with local ICP refinement to globally optimize the point-to-point registration objective. TEASER \cite{yang2020teaser} reformulates the registration problem using a Truncated Least Squares cost.

These methods provide general tools for geometric alignment, while sparse measurements and partial overlapping lead to additional challenges in intra-operative bone registration in orthopedic surgery \cite{du2026robust}. Srivatsan \emph{et al.} \cite{arun2019registration} proposed deterministic and probabilistic SPNR methods requiring intra-operative point positions and surface normals for bone registration. Min \emph{et al.} \cite{min2020generalized} combined tangents estimated from intra-operative points with pre-operative surface normals. These approaches improve registration by extracting additional geometric constraints from acquired measurements. ActiveReg instead guides probing using the current registration estimate and updates it with each new measurement, coupling registration and acquisition in a closed loop.

\subsection{Active Acquisition and Registration Assessment}

Information-driven action planning has been widely studied in active simultaneous localization and mapping \cite{placed2023survey}. In registration, Gelfand et al. \cite{gelfand2003geometrically} selected points from existing surface data to constrain unstable transformation directions in ICP. For online acquisition, Murali et al. \cite{murali2021active} selected touch actions based on predicted information gain and sequentially updated the object pose using acquired measurements. However, those methods were not designed for bone registration. ActiveReg uses the current registration estimate to plan informative region for next probing and use the actual probe measurements to update bone registration.   Fitzpatrick et al. \cite{fitzpatrick1998predicting} showed that, in point-based registration, a small fiducial registration error can coexist with a large target registration error under an unfavorable fiducial configuration. It motivates evaluating uncertainty at task-relevant locations. ActiveReg combines this assessment with pre-update innovation consistency to guide acquisition completion.

\section{Methodology}
\label{sec:methodology}

ActiveReg couples bone registration, regional probing guidance, and online completion assessment in a closed-loop framework (Fig. \ref{fig:Framework}). Each actual acquired point updates the registration estimate and geometric information. The updates inform completion assessment and guide the next recommended region when further probing is needed.

\subsection{Problem Statement}

Let $\mathcal{M}\subset\mathbb{R}^{3}$ denote the pre-operative bone surface reconstructed from CT, represented by a pre-computed SDF $D:\mathbb{R}^{3}\rightarrow\mathbb{R}$. During intra-operative probing, a tracked probe acquires contact points $\mathcal{P}_{K}=\{\mathbf{p}_{k}\}_{k=1}^{K}$ in the patient frame. The rigid transformation from the patient frame to the pre-operative model frame is represented by $\mathbf{T}=(\mathbf{R},\mathbf{t})\in SE(3)$, where $\mathbf{R}\in SO(3)$ and $\mathbf{t}\in\mathbb{R}^{3}$. The $k$-th contact point expressed in the model frame is
\begin{equation}
    \mathbf{q}_{k}(\mathbf{T}) = \mathbf{R}\mathbf{p}_{k}+\mathbf{t}.
    \label{eq:transformed_contact}
\end{equation}
The registration estimate is obtained by minimizing the SDF residuals of the accumulated contacts:
\begin{equation}
    \widehat{\mathbf{T}}_{K}
    =
    \underset{\mathbf{T}}{\arg\min}
    \sum_{k=1}^{K}
    \rho\!\left(
        D(\mathbf{R}\mathbf{p}_{k}+\mathbf{t})
    \right),
    \label{eq:robust_sdf_registration}
\end{equation}
where $\rho(\cdot)$ is a robust loss. Each residual measures the signed distance of a transformed contact to the model surface, so no predefined point correspondences are required.

Let $\mathcal{E}\subseteq\mathcal{M}$ denote the bone surface accessible under surgical exposure, assumed known for planning. \textit{Our goal is to achieve the required registration accuracy with as few acquired points as possible by coupling registration and acquisition.} The current registration estimate and its uncertainty guide where to probe next, while each newly acquired point updates the registration and informs subsequent acquisition decisions. When further probing is needed, the strategy recommends a connected probing region $\mathcal{R}_{K+1}\subseteq\mathcal{E}$ with an area of at least $A_{\min}$. The surgeon is free to choose the exact contact location within the region. The strategy also need to determine when the accumulated measurements are sufficient to end acquisition.

\subsection{Registration and Geometric Information}
ActiveReg uses the geometric information of the SDF objective to guide acquisition. After $K$ contacts have been acquired, the current registration estimate $\hat {\mathbf{T}}_K=(\hat{\mathbf{R}}_K,\hat{\mathbf{t}}_K)$ is obtained by solving Equation (\ref{eq:robust_sdf_registration}). Let $e_k = D(\mathbf{q}_k)$ denote the contact residual. Using transformation in Equation (\ref{eq:transformed_contact}), a left-multiplicative rotational perturbation and an additive translational perturbation, the residual Jacobian is
\begin{equation}
    \mathbf{J}_k = {\nabla D(\mathbf{q}_k)}^{T}
    \begin{bmatrix}
        -[\hat{\mathbf{R}}_K\mathbf{p}_k]_{\times} & \mathbf{I}_3
    \end{bmatrix},
    \label{eq:registration_jacobian}
\end{equation}
where ${[.]}_{\times}$ denotes the skew-symmetric cross-product matrix. The first three columns correspond to rotational increments and the remaining three to translational increments. Each contact provides a scalar constraint determined by its position and local SDF gradient. Complementary contacts can therefore constrain pose directions that remain weakly observed.


The robust loss $\rho$ chosen as the Cauchy loss, yielding the iteratively reweighted least squares weight is $w_k = 1 / (1+\left(e_k/c\right)^2)$. Here $c$ controls the residual scale for downweighting. The weighted Gauss–Newton matrix is
\begin{equation}
    \mathbf{H}^{\mathrm{geo}}_K
    =
    \sum_{k=1}^{K}
    w_k
    \mathbf{J}_k^{T}\mathbf{J}_k.
    \label{eq:geometric_information}
\end{equation}
Residuals, weights, and Jacobians are recomputed for all acquired contacts at each updated estimates $\hat {\mathbf{T}}_K$. Thus, Equation (\ref{eq:geometric_information}) summarizes the geometric constraints provided by the actual measurements under the current alignment. Assuming independent scalar surface measurement errors with effective variance \(\sigma_D^2\), the approximate registration information matrix and the pose covariance are
\begin{equation}
    \boldsymbol{\Lambda}^{\mathrm{reg}}_K
    =
    \frac{1}{\sigma_D^2}
    \mathbf{H}^{\mathrm{geo}}_K,
    \qquad
    \boldsymbol{\Sigma}_K
    \approx
    \left(
        \boldsymbol{\Lambda}^{\mathrm{reg}}_K
    \right)^{-1},
    \label{eq:cor}
\end{equation}
where the covariance approximation applies when $\boldsymbol{\Lambda}^{\mathrm{reg}}_K$ is positive definite. The information matrix is used to predict the information gain of candidate contacts for regional planning, while the covariance is propagated to task-relevant locations for completion assessment.

\subsection{Geometry-Guided Bootstrap Initialization}
Registration based on SDF requires an initial value, while information-driven planning requires geometric information from acquired contacts. ActiveReg establishes both through a geometry guided bootstrap using six probe contacts.

Candidate regions grow from mesh vertices designated as their centres, with geodesic-distance and surface-normal-angle limits controlling each region’s extent within the accessible surface $\mathcal E$. The first centre is closest to the mean position of the exposed vertices. Five additional centres are selected by geodesic farthest-point sampling to promote spatial coverage. The surgeon acquires one contact within each selected region. Point-to-point registration using approximate  correspondences between the six actual contacts and their region centres provides a coarse alignment. This alignment serves as the initial estimate for minimizing the SDF residuals in Equation (\ref{eq:robust_sdf_registration}). The initial information matrix $\boldsymbol\Lambda_6^{\mathrm{reg}}$ and covariance $\boldsymbol\Sigma_6$ are computed at the refined estimate $\hat{\mathbf T}_6$ using Equations (\ref{eq:geometric_information}) – (\ref{eq:cor}). The regional planner then uses $\hat{\mathbf T}_6$ and $\boldsymbol\Lambda_6^{\mathrm{reg}}$ to recommend the next probing region.

\subsection{Information-Driven Regional Acquisition Planning}

To identify informative probing regions, ActiveReg evaluates surface locations using D-optimality criterion \cite{chen2020active}. The resulting information field guides the construction of a connected region subject to a minimum-area requirement.

\subsubsection{D-optimal information gain}
For a candidate location $\mathbf{q}\in \mathcal E$, (\ref{eq:transformed_contact}) gives $\hat{\mathbf{R}}_K\mathbf{p}=\mathbf{q}-\hat{\mathbf{t}}_K$. Substituting into (\ref{eq:registration_jacobian}) yields the predicted residual Jacobian is 
\begin{equation}
    \mathbf{J}_K(\mathbf{q}) ={\nabla D(\mathbf{q})}^{T}
    \begin{bmatrix}
        -[\mathbf{q}-\hat{\mathbf{t}}_K]_{\times} & \mathbf{I}_3
    \end{bmatrix},
    \label{eq:candidate_jacobian}
\end{equation}
Assuming the inlier contact with unit robust weight and measurement variance \(\sigma_D^2\), its predicted D-optimal information gain is
\begin{equation}
    g_K(\mathbf{q}) = \log\det
    \left(
        \boldsymbol{\Lambda}^{\mathrm{reg}}_K +
        \frac{
            \mathbf{J}_K(\mathbf{q})^{T}
            \mathbf{J}_K(\mathbf{q})
        }{\sigma_D^2}
    \right)
    -
    \log\det
    \left(
        \boldsymbol{\Lambda}^{\mathrm{reg}}_K
    \right).
    \label{eq:doptimal_gain}
\end{equation}
For positive-definite $\boldsymbol{\Lambda}^{\mathrm{reg}}_K$, the matrix determinant lemma gives
\begin{equation}
    g_K(\mathbf{q})
    =
    \log
    \left(
        1+
        \frac{
            \mathbf{J}_K(\mathbf{q})
            \left(
                \boldsymbol{\Lambda}^{\mathrm{reg}}_K
            \right)^{-1}
            \mathbf{J}_K(\mathbf{q})^{T}
        }{\sigma_D^2}
    \right).
    \label{eq:doptimal_gain_compact}
\end{equation}
The gain is large when a candidate contact constrains pose directions that remain uncertain. Evaluating $g_K(\mathbf{q})$ at the accessible mesh vertices produces an information field used for region construction.


\subsubsection{Connected region construction}
To allow flexible contact placement, the pointwise information field is converted into a connected surface region with area at least $A_{\min}$. Let $F_{\mathcal{E}}$ denote the accessible triangular faces, and let $\bar{g}_K(f)$ be the mean information gain at at the three vertices of face $f$. For a threshold $\tau$, define
\begin{equation}
    F_K(\tau)
    =
    \left\{
        f\in F_{\mathcal{E}}
        \mid
        \bar{g}_K(f)\geq\tau
    \right\}.
    \label{eq:information_superlevel}
\end{equation}
Faces are connected when they share an edge. Raising $\tau$ retains higher-gain faces but may leave insufficient connected area for probing. The planner therefore selects the highest threshold admitting a connected region at least $A_{\min}$.
\begin{equation}
    \tau_K^{*}
    =
    \max
    \left\{
        \tau
        \mid
        \exists\, \mathcal{S} \in\mathcal{C}\!\left(F_K(\tau)\right),
        \;
        A(\mathcal{S})\geq A_{\min}
    \right\}.
    \label{eq:regional_threshold}
\end{equation}
Here, $\mathcal{C}(\cdot)$ denotes the connected components, $\mathcal{S}$ denotes an individual component, and $A(\mathcal{S})=\sum_{f\in \mathcal{S}}A(f)$ sums their constituent face areas. Among the components satisfying the area requirement at  $\tau_K^{*}$, the planner selects the one with the highest area-weighted mean information gain:
\begin{equation}
    \mathcal R_{K+1}
    =
    \argmax_{\substack{
        \mathcal{S}\in\mathcal{C}\!\left(F_K(\tau_K^{*})\right)\\
        A(\mathcal{S})\geq A_{\min}
    }}
    \frac{
        \sum_{f\in \mathcal{S}} A(f)\,\bar{g}_K(f)
    }{
        A(\mathcal{S})
    }.
    \label{eq:regional_selection}
\end{equation}
Area weighting accounts for unequal mesh face sizes. After a contact is acquired within $\mathcal R_{K+1}$, completion assessment determines whether probing should continue. If so, the next region is planned using the registration state updated from the actual contact.

\subsection{Online Completion Assessment}
ActiveReg combines pre-update innovation consistency with task-space uncertainty to assess completion. The former checks whether new contacts agree with preceding registration estimates, and the latter assesses if the accumulated measurements sufficiently constrain task-relevant locations.

\subsubsection{Pre-update innovation consistency}
Before updating registration with a new contact $\mathbf{p}_{K+1} $, its SDF innovation is
\begin{equation}
    \nu_{K+1}
    =
    D\!\left(
        \hat{\mathbf{R}}_{K}\mathbf{p}_{K+1} + \hat{\mathbf{t}}_{K}
    \right),
    \label{eq:innovation}
\end{equation}
The predicted innovation variance accounts for pose uncertainty and measurement noise:
\begin{equation}
    S_{K+1}
    =
    \mathbf{J}_{K+1}^{-}
    \boldsymbol{\Sigma}_{K}
    \left(\mathbf{J}_{K+1}^{-}\right)^{T}
    +
    \sigma_D^2,
    \label{eq:innovation_variance}
\end{equation}
where $\mathbf{J}_{K+1}^{-}$ is the corresponding pre-update residual Jacobian. Each innovation and its predicted variance are stored at acquisition. After \(K\) contacts, the normalized innovation statistic over the last \(W\) measurements is
\begin{equation}
    J_{\mathrm{NIS},K}
    =
    \sum_{j=K-W+1}^{K}
    \frac{\nu_j^2}{S_j}.
    \label{eq:nis}
\end{equation}
The innovation consistency condition is
\begin{equation}
    J_{\mathrm{NIS},K}
    \leq
    \chi^2_{W,\,1-\alpha},
    \label{eq:nis_condition}
\end{equation}
where $\chi^2_{W,\,1-\alpha}$ denotes the $(1-\alpha)$ quantile chi-squared distribution with \(W\) degrees of freedom. This threshold uses an approximate independent, zero-mean Gaussian innovation model.

\subsubsection{Task-space uncertainty}
Passing the innovation consistency check does not ensure that all pose directions are sufficiently constrained \cite{fitzpatrick1998predicting}. We therefore evaluate positional uncertainty at locations relevant to the surgical tasks. Let $X_{\mathrm{task}} = \left\{ \mathbf{x}_i \right\}_{i=1}^{N_T}, \mathbf{x}_i\in\mathbb{R}^{3}$ denote these locations in the pre-operative model, which may correspond to anatomical landmarks, planned resection locations, or implant-related targets at which registration error directly affects surgical execution. They are used only for uncertainty assessment and require no additional intra-operative probing.

Under the perturbation convention, the displacement sensitivity at $\mathbf{x}_i$, expressed in the model frame is
\begin{equation} 
\mathbf{G}_{i,K} = \begin{bmatrix} -[\mathbf{x}_i-\hat{\mathbf{t}}_K]_{\times} & \mathbf{I}_3 \end{bmatrix}. \label{eq:task_jacobian} 
\end{equation} 
The corresponding displacement covariance is 
\begin{equation} 
\boldsymbol{\Sigma}^{x}_{i,K} = \mathbf{G}_{i,K} \boldsymbol{\Sigma}_K \mathbf{G}_{i,K}^{T}, \label{eq:task_covariance} 
\end{equation} 
with scalar uncertainty 
\begin{equation} 
u_{i,K} = \sqrt{ \operatorname{tr} \left( \boldsymbol{\Sigma}^{x}_{i,K} \right) }. \label{eq:task_uncertainty} 
\end{equation} 
Under the local zero-mean uncertainty model, $u_{i,K}$ is the predicted Root Mean Square (RMS) displacement uncertainty at $i$-th task target. Given an uncertainty tolerance $d_{i,\mathrm{tol}}$ for each target, define
\begin{equation} 
\eta_K = \max_i \frac{ u_{i,K} }{ d_{i,\mathrm{tol}} }. \label{eq:task_uncertainty_indicator} 
\end{equation} 
The task-space uncertainty condition $\eta_K\leq1$ requires all selected task targets to satisfy their assigned tolerances.

\subsubsection{Joint Completion Criterion}
Acquisition is declared complete when $K \geq K_{\min}$, the innovation consistency condition in (\ref{eq:nis_condition}) holds, and $\eta_K \leq 1$:
\begin{equation}
    C_K
    =
    \left(K \geq K_{\min}\right)
    \land
    \left(
        J_{\mathrm{NIS},K}
        \leq
        \chi^2_{W,\,1-\alpha}
    \right)
    \land
    \left(
        \eta_K \leq 1
    \right).
    \label{eq:completion_rule}
\end{equation}
Since innovations begin after the six-contact bootstrap initialization, $K_{\min}\geq 6+W$ ensures that a full innovation window is available. If the joint criterion is not satisfied, acquisition continues until completion or the contact budget \(K_{\max}\) is reached. Reaching the budget without satisfying \(C_K\) is recorded as budget exhaustion rather than completion.

\section{Experiments and Results}
\subsection{Datasets and Experimental Setup}

\begin{table*}[t]
\centering
\caption{Registration accuracy and contact counts across bone datasets}
\label{tab:workflow}

\small
\setlength{\tabcolsep}{3pt}
\renewcommand{\arraystretch}{1.12}

\begin{adjustbox}{max width=\textwidth}
\begin{tabular}{@{}l*{4}{ccc}@{}}
\toprule
\multirow{2}{*}{Method}
& \multicolumn{3}{c}{Pelvis}
& \multicolumn{3}{c}{Proximal femur}
& \multicolumn{3}{c}{Distal femur}
& \multicolumn{3}{c}{Proximal tibia} \\
\cmidrule(lr){2-4}
\cmidrule(lr){5-7}
\cmidrule(lr){8-10}
\cmidrule(lr){11-13}
& \shortstack{MAE (R/$^\circ$)}
& \shortstack{MAE (t/mm)}
& \shortstack{$N$}
& \shortstack{MAE (R/$^\circ$)}
& \shortstack{MAE (t/mm)}
& \shortstack{$N$}
& \shortstack{MAE (R/$^\circ$)}
& \shortstack{MAE (t/mm)}
& \shortstack{$N$}
& \shortstack{MAE (R/$^\circ$)}
& \shortstack{MAE (t/mm)}
& \shortstack{$N$} \\
\midrule

\multirow{3}{*}{ICP \cite{besl1992method}}
& 32.51 & 12.45 & \textcolor{blue}{\textbf{30}}
& 21.48 & 93.23 & \textcolor{blue}{\textbf{30}}
& 41.34 &61.72  & \textcolor{blue}{\textbf{30}}
& 54.21 & 81.76 & \textcolor{blue}{\textbf{30}} \\
& 31.09 & 9.14 & 100
& 30.61 & 70.33 & 100
& 28.30 & 39.01 & 100
& 41.48 & 72.81 & 100 \\
& 26.31 & 17.54 & 200
& 23.48 & 34.21 & 200
& 31.65 & 41.37 & 200
& 19.66 & 38.52 & 200 \\
\midrule
\multirow{3}{*}{CPD \cite{myronenko2010point}}
& 21.67 & 19.85 & \textcolor{blue}{\textbf{30}}
& 30.69 & 76.24 & \textcolor{blue}{\textbf{30}}
& 45.11 & 91.27 & \textcolor{blue}{\textbf{30}}
& 59.83 & 72.08 & \textcolor{blue}{\textbf{30}} \\
& 21.37 & 16.03 & 100
& 22.47 & 31.52 & 100
& 36.91 & 34.01 & 100
& 37.04 & 54.80 & 100 \\
& 28.24 & 14.91 & 200
& 24.33 & 27.47 & 200
& 37.57 & 31.44 & 200
& 41.24 & 29.63 & 200 \\

\midrule
\multirow{3}{*}{Go-ICP \cite{yang2015go}}
& 3.41 & 2.09 & \textcolor{blue}{\textbf{30}}
& 2.87 & 4.31 & \textcolor{blue}{\textbf{30}}
& 2.22 & 5.03 & \textcolor{blue}{\textbf{30}}
& 1.96 & 3.02 & \textcolor{blue}{\textbf{30}} \\
& 3.14 & 1.32 & 100
& 2.41 & 1.27 & 100
& 2.39 & 3.18 & 100
& 1.64 & 1.52 & 100 \\
& 2.03 & 0.92 & 200
& 2.37 & 1.33 & 200
& 1.31 & 1.86 & 200
& 0.91 & 1.24 & 200 \\


\midrule
\multirow{3}{*}{Gradient-SDF \cite{li2025partial}}
& 0.86 & 0.95 & \textcolor{blue}{\textbf{30}}
& 4.26 & 4.70 & \textcolor{blue}{\textbf{30}}
& 1.58 & 1.36 & \textcolor{blue}{\textbf{30}}
& 2.99 & 3.35 & \textcolor{blue}{\textbf{30}} \\
& 0.64 & 0.73 & 100
& 1.35 & 1.51 & 100
& 1.52 & 1.06 & 100
& 1.41 & 1.91 & 100 \\
& \textcolor{red}{\textbf{0.24}}
& \textcolor{red}{\textbf{0.27}} & 200
& \textcolor{blue}{\textbf{0.83}}
& \textcolor{red}{\textbf{0.94}} & 200
& \textcolor{blue}{\textbf{0.46}}
& \textcolor{red}{\textbf{0.52}} & 200
& \textcolor{red}{\textbf{0.32}}
& \textcolor{red}{\textbf{0.62}} & 200 \\

\midrule
ActiveReg
& \textcolor{blue}{\textbf{0.62}}
& \textcolor{blue}{\textbf{0.68}}
& \textcolor{red}{\textbf{13.1}}
& \textcolor{red}{\textbf{0.58}}
& \textcolor{blue}{\textbf{1.08}}
& \textcolor{red}{\textbf{28.5}}
& \textcolor{red}{\textbf{0.38}}
& \textcolor{blue}{\textbf{1.02}}
& \textcolor{red}{\textbf{26.4}}
& \textcolor{blue}{\textbf{0.47}}
& \textcolor{blue}{\textbf{0.74}}
& \textcolor{red}{\textbf{24.3}} \\
\bottomrule
\end{tabular}
\end{adjustbox}

\par\vspace{2pt}
\begin{minipage}{\textwidth}
\footnotesize
\raggedright
\textcolor{red}{\textbf{Red}} and
\textcolor{blue}{\textbf{blue}} indicate the best and
second-best available results, respectively.
$N$: number of contacts used for registration.
\end{minipage}

\end{table*}

\begin{figure}[tpb]
	\centering
	\includegraphics[width=1\linewidth]{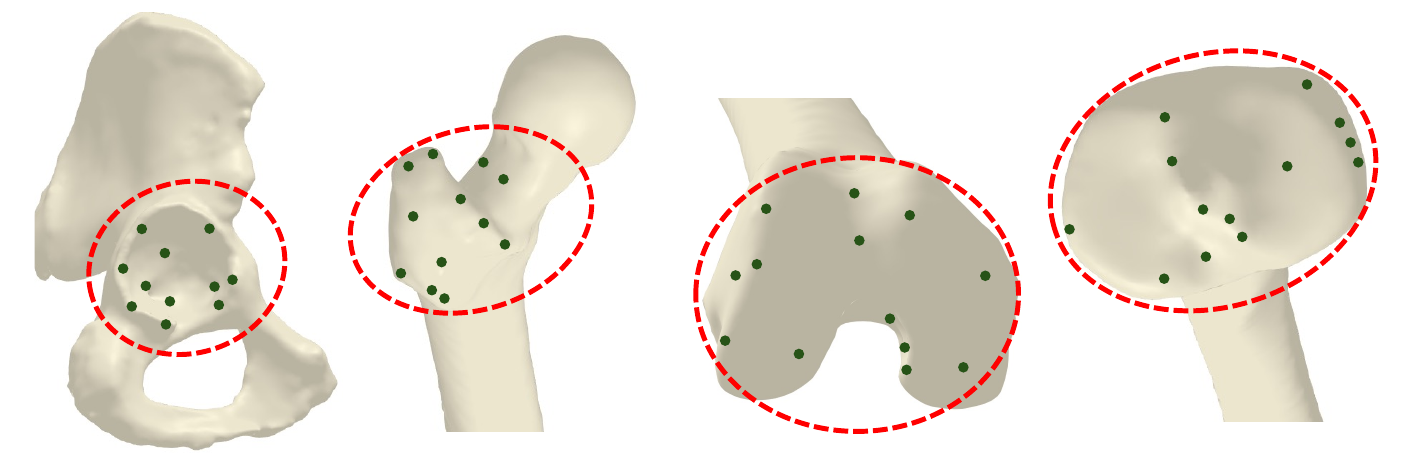}
	\caption{Anatomical exposure configurations: acetabulum (pelvis), proximal femur, distal femur, and proximal tibia. Green points indicate simulated intra-operative contacts.} 
	\label{fig:partial}
\end{figure}

In this study, the bone surface models were obtained from two publicly available datasets. RWTHmediTEC \cite{fischer2023database} provided 37 femoral and 40 hip bone models, while SSM-Tibia \cite{keast2023geometric} provided 30 right tibial models. As illustrated in Fig. \ref{fig:partial}, four exposure scenarios were defined to represent accessible surfaces in hip and knee arthroplasty, including acetabular, proximal femoral, distal femoral, and proximal tibial exposure. Each hip bone and tibia contributed one exposure scenario, while each femur contributed proximal and distal scenario, yielding 144 model–exposure configurations.

To simulate the intra-operative measurements, each bone model was transformed into the patient frame using a random rigid transformation, with rotation angles ranging from \(-45^\circ\) to \(45^\circ\) about each axis, and translation components ranging from \(-1000\,\mathrm{mm}\) to \(1000\,\mathrm{mm}\). Probe contacts were acquired within the designated exposure on the transformed model and registered to the original full pre-operative model. To simulate measurement noise during acquisition, zero-mean Gaussian noise with a standard deviation of $0.5 \mathrm{mm}$ along each coordinate axis was added to the probe contacts. The inverse of the applied transformation served as the ground truth of patient-to-model transformation. For each trial, translation and rotation errors were  calculated as the mean absolute errors across the three translation components and Euler angles, respectively.







\subsection{Closed-Loop Registration Performance}
We compared the complete ActiveReg closed-loop framework with several established registration methods, including ICP \cite{besl1992method}, CPD \cite{myronenko2010point}, Go-ICP \cite{yang2015go}, and Gradient-SDF bone registration \cite{li2025partial}. All baselines were evaluated with 30, 100, and 200 points sampled along curves on the exposed bone surface to simulate intra-operative probe motion \cite{li2025partial}. Within each trial, methods with the same contact budget used identical point sets. The 30-contact setting matches ActiveReg’s per-attempt contact budget, while the larger budgets test whether ActiveReg remains competitive against baselines using more points. ActiveReg was run with online completion assessment enabled. Attempts that did not satisfy the completion criterion within 30 contacts were restarted from bootstrap, discarding the previous measurements. Registration error was evaluated at the final output, and ActiveReg’s reported mean point count included all points acquired across attempts, including retries.

Table \ref{tab:workflow} compares registration accuracy and contact counts across four exposure scenarios. ICP and CPD always produced large errors, which may reflect the limited surface coverage and geometric ambiguity of partial-to-full registration. Across all four scenarios, ActiveReg achieved lower mean rotation and translation errors than Gradient-SDF with 30 and 100 contacts and all other baselines with 30, 100, and 200 contacts, using only 13.1–28.5 contacts on average, including retries. Across all methods and budgets, ActiveReg achieved the lowest rotation errors in both femoral scenarios and the second lowest errors in the remaining measures. Although 200-contact Gradient-SDF slightly improved six error measures, it required substantially more contacts. These results demonstrate that ActiveReg combines competitive registration accuracy with markedly reduced probing requirements, which may ease the surgeon's acquisition task intra-operatively. 


\begin{figure*}[t]
	\centering
	\includegraphics[width=1\linewidth]{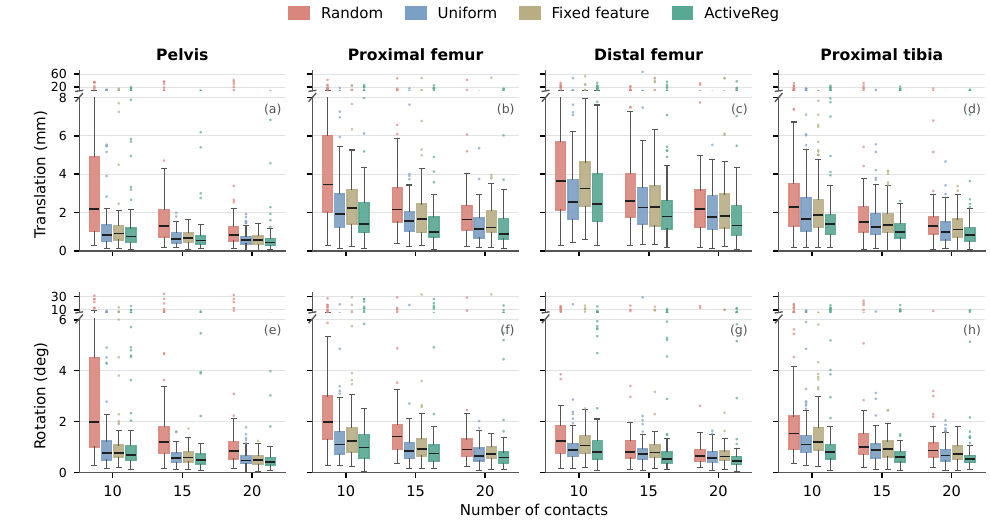}
	\caption{Registration errors under fixed contact budgets.} 
	\label{fig:4-compare}
\end{figure*}

\subsection{Effect of Acquisition Planning}
ActiveReg plans each probing region using the registration information from previously acquired points, aiming to obtain complementary constraints. To assess whether this planning helps achieve accurate registration with fewer points, we compared ActiveReg with other straightforward planning strategies that prescribe points on the exposed surface for probing. The strategies include random, geodesically uniform, and feature-guided sampling and use the same SDF registration backend.

Random sampling was uniform over the exposed surface area. Geodesically uniform sampling followed a fixed geodesic farthest-point sequence, while feature-guided sampling used surface normal variation with spatial separation. These three strategies selected sampling locations based only on the exposed surface, without using registration feedback. ActiveReg instead updated its recommended region using the registration information, with each simulated probe contact sampled uniformly over the region. Each model–exposure configuration underwent five paired trials, yielding 720 trials per strategy. Within each paired trial, all strategies shared the ground truth transformation and initial pose. For fair comparison, ActiveReg replaced the estimate from bootstrap coarse alignment with the same initial value. The registration began after six points, and all budgets included these points. The online completion assessment was disabled.

Fig. \ref{fig:4-compare} compares registration errors at budgets of 10, 15, and 20 contacts. Geodesically uniform and feature guided sampling generally reduced median errors relative to random sampling, showing the benefit of spatial coverage and consideration of surface characteristics. ActiveReg further achieved the lowest median rotation and translation errors across all four scenarios at every budget. This advantage can be attributed to its use of both surface geometry and the current registration uncertainty to recommend  regions that provide complementary constraints.

With completion assessment disabled, the fixed-budget results in Fig. \ref{fig:4-compare} also exhibits occasional large errors across all strategies. Such errors may arise when the registration converges to a local minimum corresponding to a large pose error. Information-driven planning based on this extreme incorrect estimate may not provide the measurements needed to recover an accurate alignment. These results show that acquisition planning alone does not ensure accurate registration at a fixed point budget. They highlight the role of online completion assessment in the closed-loop ActiveReg workflow and its decision reliability is also examined separately in Section \ref{study}.

\begin{table}[bhtp]
    \centering
    \caption{Target attainment rates and mean contact counts.}
    \label{tab:completion_results}
    \footnotesize
    \setlength{\tabcolsep}{1pt}
    \renewcommand{\arraystretch}{1.15}

    \begin{tabular*}{\columnwidth}
        {@{\extracolsep{\fill}}lcccccccc@{}}
        \toprule
        \multirow{2}{*}{Method}
        & \multicolumn{2}{c}{Pelvis}
        & \multicolumn{2}{c}{Prox.\ femur}
        & \multicolumn{2}{c}{Dist.\ femur}
        & \multicolumn{2}{c}{Prox.\ tibia} \\
        \cmidrule(lr){2-3}
        \cmidrule(lr){4-5}
        \cmidrule(lr){6-7}
        \cmidrule(l){8-9}
        & Rate & $\bar{K}$
        & Rate & $\bar{K}$
        & Rate & $\bar{K}$
        & Rate & $\bar{K}$ \\
        \midrule

        Random
        & 121/200 & 14.56
        & 50/185  & 16.04
        & 57/185  & 16.11
        & 62/150 & 15.08 \\

        \addlinespace[2pt]
        Uniform
        &\textcolor{red}{\textbf{190/200}}& \textcolor{blue}{\textbf{11.33}}
        & \textcolor{blue}{\textbf{106/185}} & 14.72
        & \textcolor{blue}{\textbf{61/185}} & 15.35
        & \textcolor{blue}{\textbf{89/150}} & \textcolor{blue}{\textbf{13.89}} \\

        \addlinespace[2pt]
        Feature
        & 182/200 & 11.57
        & 79/185 & \textcolor{blue}{\textbf{14.26}}
        & 48/185 & \textcolor{red}{\textbf{13.54}}
        & 75/150 & 14.01 \\

        \addlinespace[2pt]
        ActiveReg
        & \textcolor{blue}{\textbf{185/200}} & \textcolor{red}{\textbf{10.02}}
        & \textcolor{red}{\textbf{126/185}} & \textcolor{red}{\textbf{13.62}}
        & \textcolor{red}{\textbf{85/185}} & \textcolor{blue}{\textbf{14.21}}
        & \textcolor{red}{\textbf{111/150}} & \textcolor{red}{\textbf{12.27}} \\

        \bottomrule
    \end{tabular*}

    \par\vspace{2pt}
\begin{minipage}{\textwidth}
\footnotesize
\raggedright
\textcolor{red}{\textbf{Red}} and
\textcolor{blue}{\textbf{blue}} indicate the best and second best results, respectively.
\end{minipage}
\end{table}

To evaluated whether the accuracy advantage translated into fewer acquired points, we further analyse the 20-contacts trials. For each trial, we identified the first pair of consecutive acquisition steps at which both rotation and translation errors were below $1^\circ$ and $1\,\mathrm{mm}$, respectively, and recorded the contact count at the second step. Table \ref{tab:completion_results} reports the number of trials meeting this criterion out of the total (Rate) and their mean contact count (\(\bar K\)). ActiveReg achieved the highest attainment rates in both femoral scenarios and the proximal tibia. It also required the fewest contacts in the proximal femur, proximal tibia, and pelvis, demonstrating improved target attainment alongside reduced sampling requirements. Together, these results show that the proposed active planning makes more effective use of a limited contact budget, enabling the target accuracy to be reached with fewer contacts



\subsection{Ablation Studies}
\label{study}
\subsubsection{Pointwise vs Regional Probing} 
We next examined whether allowing flexible contact placement within recommended regions affected registration accuracy. ActiveReg was compared with Point-D-optimal probing, which selected the surface point with the highest predicted information gain. Both variants shared the same six bootstrap contacts and resulting coarse initialization, and used the same backend. Subsequent contacts were acquired at the planned point for Point-D-optimal and uniformly over the recommended region for ActiveReg. Five paired trials were performed per model–exposure configuration, with online completion assessment disabled and a fixed budget of 20 contacts, including bootstrap. 

Table \ref{tab:point_vs_region} shows that ActiveReg retained similar median accuracy to pointwise probing. Across the four scenarios, the largest increases in median rotation and translation errors were \(0.20^\circ\) and \(0.21\,\mathrm{mm}\), respectively, while providing recommended regions of 180 to 188 \(\mathrm{mm}^2\) corresponding to \(2.9\%\)–\(3.3\%\) of the exposed surface. It enable to provide tolerance to contact placement variation by allowing acquisition anywhere within the region without requiring precise alignment with a point target. Neither variant consistently achieved lower P90 errors across all scenarios and both metrics. Overall, the results indicate that regional planning achieves a favourable trade-off between registration accuracy and the tolerance to contact placement variation.

\begin{table}[t]
    \centering
    \caption{Point-D-optimal vs ActiveReg
    at 20 contacts.}
    \label{tab:point_vs_region}
    \footnotesize
    \setlength{\tabcolsep}{1.5pt}
    \renewcommand{\arraystretch}{1.12}

    \begin{tabular*}{\columnwidth}
        {@{\extracolsep{\fill}}llccccc@{}}
        \toprule
        \multirow{2}{*}{Scenario}
        & \multirow{2}{*}{Method}
        & \multicolumn{1}{c}{Area}
        & \multicolumn{2}{c}{Translation (mm)}
        & \multicolumn{2}{c}{Rotation ($^\circ$)} \\
        \cmidrule(lr){4-5}
        \cmidrule(l){6-7}
        & & \multicolumn{1}{c}{(mm$^2$)}
        & Median & P90 & Median & P90 \\
        \midrule

        \multirow{2}{*}{Pelvis}
        & Point-D-optimal
        & -- & 0.40 & 1.47 & 0.35 & 1.55 \\
        & ActiveReg
        & 180 & 0.45 & 1.16 & 0.41 & 1.13 \\

        \midrule
        \multirow{2}{*}{\shortstack[l]{Proximal\\femur}}
        & Point-D-optimal
        & -- & 0.93 & 2.24 & 0.44 & 1.10 \\
        & ActiveReg
        & 180 & 1.08 & 2.25 & 0.64 & 1.21 \\

        \midrule
        \multirow{2}{*}{\shortstack[l]{Distal\\femur}}
        & Point-D-optimal
        & -- & 0.96 & 2.72 & 0.36 & 0.59 \\
        & ActiveReg
        & 188 & 1.17 & 2.93 & 0.43 & 0.69 \\

        \midrule
        \multirow{2}{*}{\shortstack[l]{Proximal\\tibia}}
        & Point-D-optimal
        & -- & 0.82 & 1.43 & 0.48 & 1.15 \\
        & ActiveReg
        & 181 & 0.79 & 1.49 & 0.55 & 0.91 \\

        \bottomrule
    \end{tabular*}
    \par\vspace{2pt}
\begin{minipage}{\textwidth}
\footnotesize
\raggedright
P90 denotes the 90th percentile of registration errors across trials.
\end{minipage}
    
\end{table}

\subsubsection{Online Completion Assessment} 
We evaluated whether the proposed online completion assessment can identify when contact acquisition is sufficient without access to ground truth. Three variants were compared within the ActiveReg framework: innovation consistency alone (NIS), task-space uncertainty alone (Uncertainty), and their conjunction (Joint). All variants used the same budget \(K_{\max}=30\). The innovation window was \(W=5\), with a significance level of \(\alpha=0.05\). Task-space uncertainty was evaluated at five locations covered each bone model as surrogate targets, since the datasets lacked annotated surgical task targets. Table \ref{tab:completion_ablation} reports the completion rate, along with the mean contact count, and the mean/maximum registration errors among completed trials. 

NIS alone declared completion after approximately 13 contacts on average, but substantial errors remained, indicating that measurement consistency alone did not establish accurate registration. Uncertainty alone improved performance in several scenarios, yet still accepted large-error estimates in the pelvis and proximal femur. Joint achieved the lowest mean and maximum rotation and translation errors across all four scenarios. Compared with Uncertainty alone, its mean contact count increased by at most 1.03, with only one to three fewer completed trials per scenario. These results support combining task-space uncertainty assessment with innovation consistency to improve the reliability of completion decisions.

\begin{table}[t]
    \centering
    \caption{Ablation of the online completion criterion.}
    \label{tab:completion_ablation}
    \footnotesize
    \setlength{\tabcolsep}{2pt}
    \renewcommand{\arraystretch}{1.12}

    \begin{tabular*}{\columnwidth}
        {@{\extracolsep{\fill}}llcccc@{}}
        \toprule
        \multirow{2}{*}{Scenario}
        & \multirow{2}{*}{Criterion}
        & Rate
        & \multirow{2}{*}{$\bar{K}$}
        & Rot.\ ($^\circ$)
        & Trans.\ (mm) \\
        & & Pass / Total
        & & MAE / Max
        & MAE / Max \\
        \midrule

        \multirow{3}{*}{Pelvis}
        & NIS         & \textbf{197 / 200} & 13.65 & 1.17 / 23.32 & 1.22 / 26.99 \\
        & Uncertainty & 196 / 200 & \textbf{13.04} & 1.54 / 21.23 & 1.74 / 27.61 \\
        & Joint       & 194 / 200 & 13.61 &  \textbf{0.71 / 1.96} & \textbf{0.75 / 1.91} \\

        \midrule
        \multirow{3}{*}{\shortstack[l]{Proximal\\femur}}
        & NIS         & \textbf{183 / 185} & \textbf{13.35} & 1.32 / 21.35 & 2.13 / 27.11 \\
        & Uncertainty & 120 / 185 & 20.22 & 1.30 / 17.20 & 1.83 / 22.56 \\
        & Joint       & 117 / 185 & 21.15 & \textbf{0.59 / 1.41} & \textbf{0.97 / 2.16} \\

        \midrule
        \multirow{3}{*}{\shortstack[l]{Distal\\femur}}
        & NIS         & \textbf{185 / 185} & \textbf{14.05} & 1.62 / 19.43 & 2.12 / 13.93 \\
        & Uncertainty & 166 / 185 & 22.75 & 0.47 / 1.98 & 1.18 / 2.42 \\
        & Joint       & 164 / 185 & 22.70 & \textbf{0.43 / 0.88} & \textbf{1.17 / 2.09} \\

        \midrule
        \multirow{3}{*}{\shortstack[l]{Proximal\\tibia}}
        & NIS         & \textbf{150 / 150} & \textbf{13.55} & 0.76 / 2.84 & 1.19 / 4.11 \\
        & Uncertainty & 149 / 150 & 20.75 & 0.58 / 1.94 & 1.24 / 3.11 \\
        & Joint       & 148 / 150 & 21.78 & \textbf{0.48 / 1.14} & \textbf{0.79 / 2.32} \\

        \bottomrule
    \end{tabular*}

    \par\vspace{2pt}
\begin{minipage}{\textwidth}
\footnotesize
\raggedright
\textbf{Bold} indicates the best performance.
\end{minipage}
\end{table}

\begin{figure}[htpb]
	\centering
	\includegraphics[width=1\linewidth]{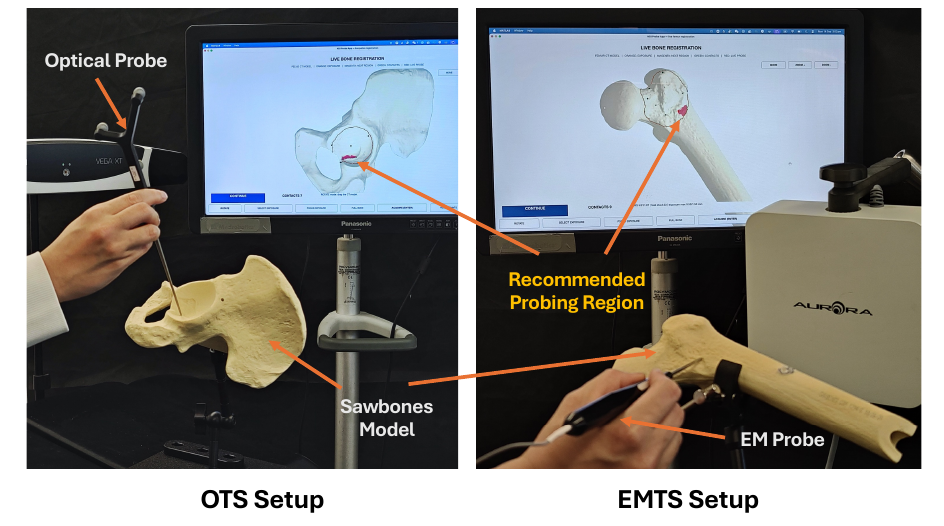}
	\caption{Phantom setups: repeated trials were performed with the sawbones models in each of the two tracking systems. ActiveReg displays recommended probing regions on the bone model to guide manual acquisition with a tracked probe.} 
	\label{fig:phantom}
\end{figure}

\subsection{Real Phantom Experiments}

ActiveReg was further evaluated on phantoms to assess registration accuracy and contact requirements during manual probing (Fig. \ref{fig:phantom}). Four pelvic and three femoral Sawbones models were scanned to generate the pre-operative bone models. Experiments were conducted with an optical tracking system (OTS) \cite{sorriento2019optical} and an electromagnetic tracking system (EMTS) \cite{li2023closed}. The OTS was a Polaris Vega XT, and the EMTS was an Aurora system with a planar field generator (both from Northern Digital Inc., Canada). The recommended regions computed by ActiveReg were highlighted on the displayed bone model to guide probing.

For the baseline methods, the operator collected the points by moving the tracked probe on the exposed bone surface and stopped when they judged its coverage sufficient. ICP, CPD, Go-ICP, GeoTransformer, and Gradient-SDF bone registration used the same acquired point set in each trial. ActiveReg instead guided probing through regional recommendations and used the online completion assessment to determine when acquisition could end. Registration accuracy was evaluated using target registration error (TRE) at ten anatomical landmarks distributed across each bone model. Their positions were acquired on the phantom with the tracked probe and identified on the corresponding scanned model. These landmarks were excluded from registration. As shown in Table \ref{tab:phantom_results}, ActiveReg achieved comparable mean TRE to Gradient-SDF with an average of only $24 – 28$ acquired points, compared with $771 – 1263$ for dense sampling. The difference in mean TRE remained below $1\,\mathrm{mm}$ across all four conditions.

\begin{table}[t]
    \centering
    \caption{Registration results in phantom experiments.}
    \label{tab:phantom_results}
    \footnotesize
    \setlength{\tabcolsep}{2pt}
    \renewcommand{\arraystretch}{1.12}

    \begin{tabular*}{\columnwidth}
        {@{\extracolsep{\fill}}llcccc@{}}
        \toprule
        \multirow{3}{*}{Setup}
        & \multirow{3}{*}{Method}
        & \multicolumn{2}{c}{Pelvis}
        & \multicolumn{2}{c}{Femur} \\
        \cmidrule(lr){3-4}
        \cmidrule(l){5-6}
        & & TRE (mm) & $\bar{K}$ & TRE (mm) & $\bar{K}$ \\
        \midrule

        \multirow{5}{*}{OTS}
        & ICP \cite{besl1992method}         &73.61 &1263.2 &90.45 &1104.5 \\
        & CPD \cite{myronenko2010point}         &67.24 &1263.2 &47.18 &1104.5 \\
        & Go-ICP \cite{yang2015go}      &9.72  &1263.2 &10.27 &1104.5 \\
        & GeoTrans \cite{qin2023geotransformer}    &8.86  &1263.2 &6.07 &1104.5 \\
        & Gradient-SDF \cite{li2025partial} &\textcolor{red}{\textbf{3.47}}  & 1263.2 &\textcolor{red}{\textbf{3.61}} &1104.5 \\
        & ActiveReg    &\textcolor{blue}{\textbf{3.97}}  &\textcolor{red}{\textbf{23.6}}   &\textcolor{blue}{\textbf{4.46}} & \textcolor{red}{\textbf{27.6}}\\

        \midrule

        \multirow{5}{*}{EMTS}
        & ICP \cite{besl1992method}         &65.13 &942.5 &112.51 &771.3 \\
        & CPD \cite{myronenko2010point}          &87.02 &942.5 &42.46  &771.3 \\
        & Go-ICP \cite{yang2015go}      &6.83  &942.5 &12.11 &771.3 \\
        & GeoTrans \cite{qin2023geotransformer}    &5.41  &942.5 &7.65 &771.3 \\
        & Gradient-SDF \cite{li2025partial} &\textcolor{red}{\textbf{3.62}}  &942.5 &\textcolor{blue}{\textbf{4.25}} &771.3 \\
        & ActiveReg    &\textcolor{blue}{\textbf{4.04}}  &\textcolor{red}{\textbf{25.1}} &\textcolor{red}{\textbf{4.12}}  &\textcolor{red}{\textbf{28.3}} \\

        \bottomrule
    \end{tabular*}

        \par\vspace{2pt}
\begin{minipage}{\textwidth}
\footnotesize
\raggedright
\textcolor{red}{\textbf{Red}} and
\textcolor{blue}{\textbf{blue}} indicate the best and second best results, respectively.
\end{minipage}
\end{table}






\section{CONCLUSIONS}
\label{sec: conclusion}

This work presents ActiveReg, a closed-loop framework for accurate bone registration with few intra-operative probe contacts. ActiveReg uses registration information from actual acquired contacts to plan the next recommended probing regions and performs online completion assessment to determine whether further probing is needed. The recommended regions provide explicit guidance on where to place the probe while allowing flexibility in the exact contact location.

Extensive simulations across four anatomical scenarios and phantom experiments with two tracking systems demonstrated that ActiveReg achieved competitive accuracy compared with baseline registration methods while using substantially fewer acquired points. The ablation studies showed that regional guidance balanced registration accuracy with tolerance to contact placement variation, while combining innovation consistency and task-space uncertainty improved the reliability of completion decisions. Future work will assess acquisition time, surgeon effort, and registration accuracy under realistic surgical conditions.










\bibliographystyle{ieeetr}
\bibliography{reference}

\end{document}